\documentclass{article}
\usepackage{arxiv}

\usepackage[utf8]{inputenc} 
\usepackage[T1]{fontenc}    
\usepackage{hyperref}       
\usepackage{url}            
\usepackage{booktabs}       
\usepackage{amsfonts}       
\usepackage{nicefrac}       
\usepackage{microtype}      
\usepackage{lipsum}		
\usepackage{graphicx}
\usepackage{natbib}
\usepackage{doi}
\usepackage{amsmath}
\usepackage{amssymb}
\usepackage{multirow}

\pdfoutput=1
\usepackage[pdftex]{}

\newcommand{\mypar}[1]{\vspace{3pt}\noindent\textbf{#1~}}
\newcommand{\zz}{\mathbf{z}}
\newcommand{\xx}{\mathbf{x}}
\newcommand{\yy}{\mathbf{y}}

\newcommand{\XX}{\mathcal{X}}
\newcommand{\YY}{\mathcal{Y}}
\newcommand{\pphi}{\boldsymbol{\phi}}
\newcommand{\mr}[1]{\mathrm{#1}}
\newcommand{\Loss}{\mathcal{L}}
\newcommand{\ppm}{\,\scriptsize$\pm$}

\title{TTTFlow: Unsupervised Test-Time Training with Normalizing Flow}

\date{} 					

\author{David Osowiechi\thanks{Equal contribution} \And Gustavo A. Vargas Hakim\footnotemark[1] \And Mehrdad Noori \And Milad Cheraghalikhani \And Ismail Ben Ayed \And Christian Desrosiers}
\date{LIVIA, ÉTS Montréal, Canada \\ International Laboratory on Learning Systems (ILLS), \\ McGILL - ETS - MILA - CNRS - Université Paris-Saclay - CentraleSupélec, Canada \texttt{david.osowiechi.1@ens.etsmtl.ca,  gustavo-adolfo.vargas-hakim.1@ens.etsmtl.ca,
mehrdad.noori.1@ens.etsmtl.ca, milad.cheraghalikhani.1@ens.etsmtl.ca\\
ismail.benayed@etsmtl.ca,  christian.desrosiers@etsmtl.ca}}

\renewcommand{\shorttitle}{TTTFlow: Unsupervised Test-Time Training with Normalizing Flow}

\hypersetup{
pdftitle={TTTFlow},
pdfsubject={cs.ML, cs.AI, cs.CV},
pdfauthor={David Osowiechi, Gustavo A. Vargas Hakim, Mehrdad Noori, Milad Cheraghalikhani, Ismail Ben Ayed, Christian Desrosiers},
pdfkeywords={Test-time Adaptation, Image Classification, Normalizing Flows, Unsupervised Training},
}

\begin{document}
\maketitle

\begin{abstract}
A major problem of deep neural networks for image classification is their vulnerability to domain changes at test-time. Recent methods have proposed to address this problem with test-time training (TTT), where a two-branch model is trained to learn a main classification task and also a self-supervised task used to perform test-time adaptation. However, these techniques require defining a proxy task specific to the target application. To tackle this limitation, we propose TTTFlow: a Y-shaped architecture using an unsupervised head based on Normalizing Flows to learn the normal distribution of latent features and detect domain shifts in test examples. At inference, keeping the unsupervised head fixed, we adapt the model to domain-shifted examples by maximizing the log likelihood of the Normalizing Flow. Our results show that our method can significantly improve the accuracy with respect to previous works.
\end{abstract}


\section{Introduction}
\label{sec:introduction}

Deep learning has become increasingly effective for computer vision tasks such as segmentation or classification. 
Nevertheless, these achievements are often made under the assumption that training and test data share the  same distribution, which is not always the case in practice. Furthermore, a small distribution shift between the training and test data can lead to an important drop in model performance \cite{Recht_2018_cifar101}. Two types of methods were proposed to increase the robustness of the model to distributional change: Domain Generalization and Domain Adaptation. Domain Generalization (DG)~\cite{dg1,dg2,dgsurvey} involves training on a large set of source data from several domains to help the model be more robust and generalize to unseen domains. However, DG requires a large amount of data from different domains, which can be difficult to obtain, and there is no guarantee that the new model can generalize well to an unseen domain at test-time. On the other hand, Domain Adaptation (DA)~\cite{da1,shot,dasurvey} aims to avoid performance degradation of a model trained on a source domain when used on a test set from a different domain. The distribution shift in this context is reduced without prior training on different domains, but in some cases requires access to the labeled source samples.

Test-Time Adaptation (TTA)~\cite{tta1,wang_tent:_2021,sun_test-time_2020,boudiaf2022} is an emerging field that studies approaches to quickly adapt a pretrained deep network to domain shifts during test-time. Unlike DG, the source training typically involves a single domain. Moreover, in contrast to DA, it is possible to fine-tune the network at test-time. This task remains challenging as it is expected that the source data is not available during test-time, hence directly measuring the domain discrepancy becomes complicated. However, finding a solution for TTA is an attractive endeavor, as it promises a more widely useful deployment of deep networks in real-world contexts. Recent TTA techniques have explored adapting batch statistics in the feature extractor of deep networks~\cite{wang_tent:_2021}. While this strategy provides some robustness, it is often suboptimal, as a sufficiently diverse batch of samples is needed in order to capture enough information to adapt the weights of the network. Another strategy, inspired by self-supervised learning, is to include additional tasks~\cite{sun_test-time_2020,liu2021ttt++}. Although this strategy has been used successfully for TTA, it is sensitive to the 
chosen proxy task and requires pseudo-labels.

In this paper, we present TTTFlow, a novel approach for unsupervised Test-Time Adaptation, which makes use of a Normalizing Flow as a domain shift detector. Our method does not require access to the source data during inference, can measure domain discrepancy in a tractable way, and does not require special proxy tasks to be solved along with the source training. Moreover, TTTFlow could be built on top of any off-the-shelf network without additional technical adjustments.\par 
Specifically, the contributions of this work can be summarized as follows:
\begin{itemize}
    \item We introduce an unsupervised method under the test-time training paradigm of TTA. Our approach is designed to  directly measure the domain shift between target and source images, without the need of an extra task.
    \item To the best of our knowledge, this is the first work that employs Normalizing Flows to measure domain shift in Test-Time Adaptation. While they have been recently investigated for domain alignment, their application in tasks related to Domain Adaptation remains unexplored.
\end{itemize}
The remainder of this paper is organized as follows. Section~\ref{sec:related} presents prior work on Test-Time Adaptation. Section~\ref{sec:method} then introduces the proposed TTTFlow method. Section~\ref{sec:experiments} describes the experimental setup used to evaluate our method, and Section~\ref{sec:results} reports the results.

\section{Related Work}
\label{sec:related}

\mypar{Normalizing Flows} In the field of generative modeling, Normalizing Flows have gained important traction due to their capabilities of learning tractable distributions in the latent space \cite{normflows,normflows2}. Their goal is to transform a generally unknown data distribution into a known one, typically the normal distribution, from which we can easily sample new data points and measure their exact likelihood. The transformation of the flow model is guaranteed to be bidirectional by using an invertible and differentiable architecture. Although Normalizing Flows have not been formally used in the field of Domain Adaptation, recent works suggest that they can be an effective tool for Domain Alignment \cite{domainalign1,domainalign2}, where the domain of two different datasets must be fitted with indistinguishable distributions. In this work, we take a step further by proposing Normalizing Flows as an alternative to learn and codify a domain, so that it can later be used at test-time.

\mypar{Test-Time Adaptation} These methods allow using off-the-shelf models without any additional training. In general terms, test-time adaptation focuses on adapting models that were not trained with a special configuration prior to being used at inference. One of the first approaches of this category, called TENT \cite{wang_tent:_2021}, requires to be given the model and target data. It then updates the model layers containing normalization statistics by minimizing the Shannon entropy of predictions. The authors of \cite{Mummadi_2021} improve TENT by using a log-likelihood ratio instead of entropy, and by estimating the statistics of the target batch. In SHOT \cite{shot}, the entire feature extractor is fine-tuned using a mutual information loss along with pseudo-labels to correct inaccurate predictions from the pretrained model. LAME \cite{boudiaf2022} is an adaptation method that do not alter the network layers, but just focuses on a post-hoc adaptation of the softmax predictions through Laplacian regularization.  

\mypar{Batch Norm Adaptation} To increase robustness of a model, Batch Normalization (BN) can be used for a faster convergence and increased stability during training. Nevertheless, a shift in the distribution causes the statistics to change, which is why some papers suggest adapting normalization statistics to improve performance. For instance, Prediction Time Batch Normalization~\cite{PTN} proposes to use the mean and  variance from the batch of test samples as statistics in the batch norm layer. However, this estimation can be inaccurate due to a small number of data samples.
To avoid this, the authors of \cite{Schneider_2020}  compute a new mean and variance, which is a mix of the BN statistics computed at training and the new estimation at test time. The same approach is used by SITA \cite{SITA} to estimate statistics, with the difference that it can be used on a single data example. To achieve this, SITA generates a pseudo-batch by randomly augmenting this example and then computes the statistics on this pseudo-batch.

\mypar{Test-Time Training} Methods based on Test-Time Training (TTT) \cite{sun_test-time_2020} update the model at inference, but use a Y-shaped architecture with a main task and a self-supervised task which are learned at training time. The model is trained by jointly minimizing the losses in both branches. After the model has been trained, the parameters of the main task branch are frozen. At test-time, the parameters of the shared encoder are updated so to minimize the self-supervised loss. Following this approach, \cite{sun_test-time_2020} uses rotation prediction~\cite{Gidaris_2018} as self-supervised task, where images are randomly rotated by multiples of 90$^{\circ}$ (0$^{\circ}$, 90$^{\circ}$, 180$^{\circ}$, 270$^{\circ}$) and the model should recover this rotation. A major problem of this approach is the choice of the self-supervised loss, which should be related to both source and target datasets. Inspired by TTT, TTT++~\cite{liu2021ttt++} adds a loss which promotes online feature alignment by comparing the statistics of the source data with those of the current batch. For the self-supervised task, the rotation prediction loss is replaced by a contrastive loss which encourages the encoded features for two different augmentations of the same image to be similar, and the ones of different images to be dissimilar. Lastly, the authors of MT3 \cite{MT3} use meta training at inference on the second task to improve the performance of TTT.

\section{Method}
\label{sec:method}

\begin{figure*}
    \centering
    \includegraphics[width=.88\linewidth]{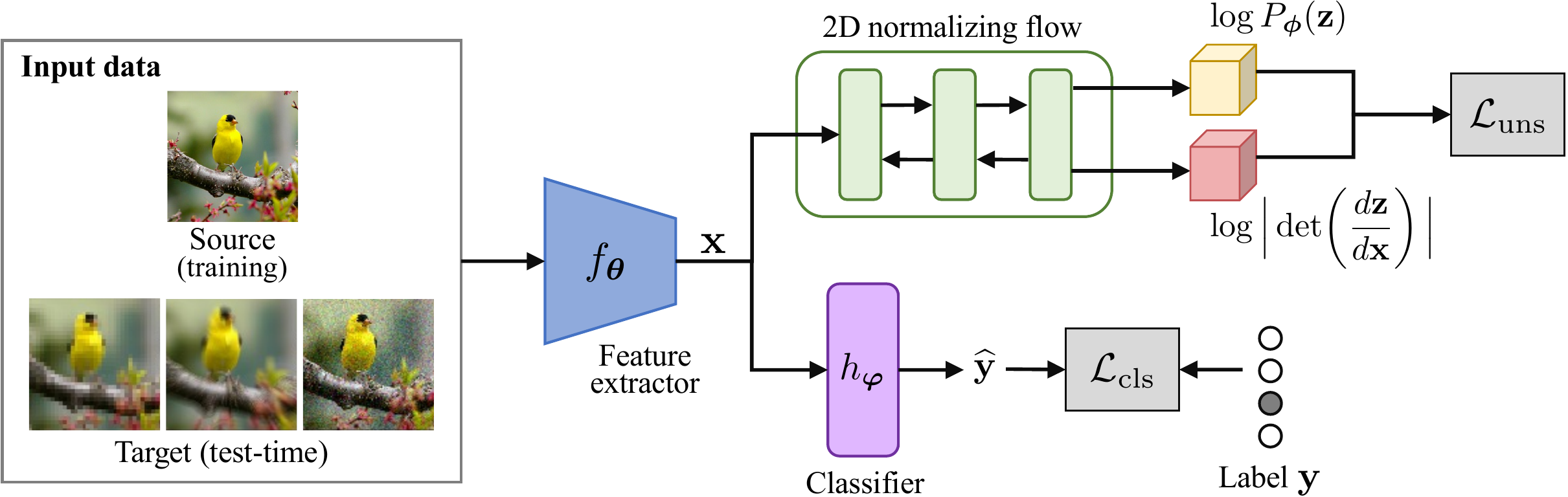}
    \caption{Architecture of TTTFlow. The feature maps from a pretrained network are transformed into an isotropic Gaussian distribution through a Normalizing Flow $g_\phi$. The log-likelihood of an input $\xx$ is measured based on the likelihood in the latent space $\zz = g_\phi (\xx)$. At test-time, the frozen unsupervised head is then used as a domain shift detector to fine-tune the extractor. Target images show examples of pixelate, zoom blur and Gaussian noise corruptions (from left to right) from the CIFAR-10-C dataset.}
    \label{fig:tttflow}
\end{figure*}

We start by defining the problem of Test-Time Training and then present our TTTFlow method for this problem.

\subsection{Problem definition}

In the context of classification, we denote the domain as the joint distribution $P_{XY}$ between the input space $\XX$ and the label space $\YY$, and define the marginal distribution of the inputs as $P_{X}$. 
At training time, a deep network learns from the data out of a source domain ($\XX_{s}, \YY_{s}$), whilst
at test-time, the network must be adapted to a new target domain ($\XX_{t}, \YY_{t}$), such that $P_{X_{s}} \neq P_{X_{t}}$. 
Both domains share the same label space ($\YY_{s} = \YY_{t}$), but the labels for the target inputs are unknown. 
The goal of Test-Time Training is to learn a function $g_{f}: \XX_{t} \rightarrow \YY_{t}$ on the basis of an already known function $f: \XX_{s} \rightarrow \YY_{s}$.

\subsection{Proposed framework}
\label{subsec:CNN}

Our framework exploits a multi-head architecture where a Normalizing Flow is used to encode domain-specific information from a pretrained feature extractor. An important benefit of this configuration is that it can be applied on any pretrained network without the need of a special training on the source data. In what follows, we describe the different components of TTTFlow and the mechanisms that allow test-time training with domain shift. The overall scheme of the model can be seen in Fig.~\ref{fig:tttflow}.

\subsection{Source Training}

Although our TTFlow method can be used on top of any architecture, in this work we consider a CNN as the classification backbone. This CNN can be divided in a feature extractor $f_{\boldsymbol{\theta}}$ (parameterized by $\boldsymbol{\theta}$) followed by a classifier head $h_{\boldsymbol{\varphi}}$ (parameterized by $\boldsymbol{\varphi}$). Let $\xx = f_{\boldsymbol{\theta}}(\textbf{I})\in\mathbb{R}^{c\times h\times w}$ be the 2D feature map from an input image $\textbf{I}$, and $\hat{\yy}=h_{\boldsymbol{\varphi}}(\xx)\!\in\![0,1]^K$ be the softmax predictions from the classifier, where $K$ is the number of classes. The source training of the network is performed in a supervised way using the cross-entropy loss. However, and as shown in further sections, using a more robust source training (e.g., adding contrastive learning) can help to achieve better adaptation results.

\subsection{Normalizing flows as a domain shift detector}
\label{subsec:NF}

Once the CNN is trained, we need a way to encode the source domain distribution, such that domain shifts can be detected at test time. We propose using Normalizing Flows \cite{normflows,realnvp,glow} for this purpose, because of their ability to model complex, high-dimensional distributions effectively. Normalizing Flows are generative models capable of transforming data from a complex and often unknown distribution into a latent space with a well-defined and tractable distribution. 
In this study, the feature map $\xx$ follows an unknown distribution $P(\xx)$, which is related to $P_{X}$. A function $g_{\pphi}$ transforms the feature map into its latent representation $\zz = g_{\pphi}(\xx)\in\mathbb{R}^{c\times h\times w}$ with $\zz\sim P_{\pphi}(\zz)$, such that $P_{\pphi}(\zz)$ is a tractable distribution (e.g. standard multi-variate Gaussian distribution) with a known probability density function. 
The flow-based function $g_{\pphi}$ should meet two requirements: (1) being invertible, i.e. $\xx = g_{\pphi}^{-1}(\zz)$, and (2) being differentiable w.r.t. the input in both directions. Furthermore, a higher representation power can be achieved if a composition of invertible and differentiable functions is used: $g_{\pphi}=g_{1}\circ g_{2}\circ \cdots g_{M}$. Knowing that $\textbf{x}$ is transformed into $\textbf{z}\sim P_{\pphi}$, the likelihood of the original variable can then be computed exactly using the change of variable rule,
\begin{align}
    \log P(\xx) & \, = \, \log P_{\pphi}(\zz) \, + \, \log \Big| \det\!\left( \frac{d\zz}{d\xx} \right)\Big| \nonumber\\ 
    & \, = \, \log P_{\pphi}(\zz) + \sum_{i=1}^{M}\log\Big|\det\! \left(\frac{d g_{i}}{dg_{i-1}} \right)\Big|,
    \label{eq:loglike}
\end{align}
where $\log|\det (dg_{i}/dg_{i-1})|$ is the logarithm of the Jacobian matrix determinant.

Affine coupling layers are a popular choice to build Normalizing Flows \cite{realnvp,glow}, so that the resulting Jacobian matrix is upper triangular and its determinant is easily computed as the product of its diagonal elements. The model can be trained by minimizing the negative log-likelihood in Eq.~(\ref{eq:lossuns}):
\begin{equation}
    \Loss_{\mr{uns}} \, = \, -\log P(\xx)
    \label{eq:lossuns}.
\end{equation}

A Normalizing Flow based on RealNVP \cite{realnvp} is placed on top of the frozen feature extractor $f_{\boldsymbol{\theta}}$ to learn the latent space of $\zz$ from the source inputs $\xx\sim\XX_{s}$ in an unsupervised way (i.e., using Eq.~(\ref{eq:loglike}) directly). We hypothesize that this model captures the domain information from the source data, thus can be used to measure domain shift in the target data. 

\subsection{Test-time training with flow-based model}

At test-time, the pretrained network must adapt its parameters to unlabeled inputs from an also unknown target domain $X_{t}$. We achieve this by only focusing on the extractor parameters $\boldsymbol{\theta}$, similarly to \cite{sun_test-time_2020,liu2021ttt++}. 
The frozen Normalizing Flow transforms each new test image feature map $\xx_{t}$ into its latent representation $\zz_{t} = g_\phi(\xx_{t})$ to compute its log-likelihood using Eq.~(\ref{eq:lossuns}). Note again that the log-likelihood is measured with respect to the multivariate Gaussian distribution, into which the unsupervised head transforms the features' distribution. This value provides information of the domain shift, as a feature map that is closer to the latent space of the source data should have a higher log-likelihood than a feature map that is farther away. Hence, negative log-likelihood can be once again used as the loss function to adapt the extractor for the target input. 

\section{Experimental setup}
\label{sec:experiments}

We evaluate our TTTFlow method on two popular test-time adaptation benchmarks based on the CIFAR-10 dataset \cite{Krizhevsky09Cifar10}, CIFAR-10-C \cite{hendrycks2019robustness_Cifar10C} and CIFAR-10.1 \cite{Recht_2018_cifar101}, and compare its performance against state-of-art approaches for this task. As explained in Section \ref{subsec:CNN}, the first step is to train a CNN on source data from CIFAR-10, a natural image classification dataset consisting of 10,000 images for training and 2,000 images for testing, with 10 different classes.

Once the CNN is trained, the training of the Normalizing Flow (NF) is performed as explained in Section \ref{subsec:NF}. We adopted RealNVP \cite{realnvp} as it has become a standard tool for flow modeling. Our compact version is made of three coupling layers with two resblocks in each of them. The \emph{checkerboard} coupling was found to be more effective than its channelwise counterpart. Similar to \cite{liu2021ttt++}, the flow model is placed on top of the second layer of the ResNet50's feature extractor based on the common assumption that domain information is mostly located at the early stages of feature extraction while class information is encoded at later stages \cite{zhou2020domain}. More implementation details can be found in the supplementary material, as well as the corresponding ablation study on the flow architecture.

We use CIFAR-10 images without any label information for this step, as the NF model is trained in an unsupervised manner. The training was performed for $100$ epochs using SGD with an initial learning rate of $0.1$ and a cosine annealing scheduler.

Following previous work, ResNet50 \cite{Resnet50} is chosen as the main architecture. The model is trained for 350 epochs with SGD, using a batch size of 128 images and an initial learning rate of $0.1$ which is reduced by a factor of 10 at epochs 150 and 250.

After the two-step source training, the NF model is used to detect domain shift through negative log-likelihood and update the part of the network from where the features are collected (i.e., up until second layer). For all the experiments at test-time, we keep the batch size of 128 images and use a learning rate of 0.001, along with SGD as the main optimizer. At each new batch, we initialize our feature extractor with the weights of the learning part. This is to avoid computing on an error made by the optimization, and is based on the assumption that each batch can have different corruptions as made by \cite{sun_test-time_2020} in their offline mode.

We use the pretrained CNN as baseline, and compare our results against TENT \cite{wang_tent:_2021}, TTT \cite{sun_test-time_2020}, and TTT++ \cite{liu2021ttt++}. 
For a fair comparison, we reproduced these previous methods under the same experimental conditions as in TTTFlow, i.e. using the same hyperparameters such as batch size, number of adaptation epochs, and so on. Our codebase can be found in \url{https://github.com/GustavoVargasHakim/TTTFlow.git}.

\section{Results and discussion}
\label{sec:results}

\begin{table}[t!]
    \centering
    \caption{Comparison of joint versus separate training for TTT and TTTFlow  on CIFAR-10-C data with Level 5 Gaussian Noise Corruption.}
    	\label{tab:ComparisonMethod}
    \begin{small}
            \begin{tabular}{ll|c}
            \toprule
    \multicolumn{2}{l|}{\bf Method} & \bf Accuracy (\%) \\ 
    \midrule
    \multirow{2}{*}{TTT~\cite{sun_test-time_2020}} & Separate & 61.18 \\
    &  JT as in \cite{sun_test-time_2020} & 57.96\\
    \midrule
    \multirow{3}{*}{TTTFlow} & Separate & 62.75 \\
    &JT ($\beta$ = 0.01) & 58.16 \\ 
    & JT ($\beta$ = 0.001) & 58.44 \\
    \bottomrule
            \end{tabular}   
            \end{small}        
\end{table}

We first perform ablation and comparison experiments on the CIFAR-10-C dataset containing different types of image corruption, and then extend our evaluation to natural domain shift using the CIFAR-10.1 dataset.

\subsection{Object recognition on corrupted images}

Our first experiments evaluate TTTFlow on the CIFAR-10-C dataset which comprises 15 different algorithmic corruptions (e.g. Gaussian noise, zoom blurring, etc.) with 10,000 images each (see Fig.~\ref{fig:tttflow} for examples). Each corruption has five severity levels, with Level 1 corresponding to mild corruptions and Level 5 to strongest ones. Unless specified otherwise, we evaluate TTTFlow on Level 5, as it represents the most challenging adaptation scenario. 

\mypar{Joint vs separate training} As a first step, we compare our method, which learns the NF on top of a frozen classifier (separate training), with the Joint Training (JT) approach training the classification task and unsupervised task at the same time by minimizing
\begin{equation}
    \Loss_{\mr{JT}} \, = \, \Loss_{\mr{cls}} + \beta  \Loss_{\mr{uns}}
    \label{eq:loss},
\end{equation}

where hyperparameter $\beta$ controls the trade-off between the two losses. This JT approach follows previous work on test-time training \cite{sun_test-time_2020,liu2021ttt++}. An important problem with this approach is the need to retrain the main classification network when learning occurs along with the secondary task. To avoid this issue and to exploit the weights of any pretrained backbone, we freeze all the parameters of the CNN, except the batch norm statistics of the feature extractor, and proceed to train the NF independently. This enables using any backbone without retraining.

Table~\ref{tab:ComparisonMethod} gives the accuracy of our method using separate training or JT with $\beta\!=\!0.01$ or $\beta\!=\!0.001$. As can be seen, placing the NF model on a pretrained encoder yields better performance than performing joint training. We conjecture that the Normalizing Flow is particularly sensitive to the joint training, as it is forced to learn a Gaussian distribution out of the continually changing feature maps distribution, as they are being modified for classification. Moreover, the information needed to encode the domain of examples seems different from the information needed to classify them. The same analysis is performed for TTT~\cite{sun_test-time_2020}, which has a rotation prediction semi-supervised loss in addition to the classification loss. As reported in Table~\ref{tab:ComparisonMethod}, we find once again that training the TTT model in two separate steps is better than the JT approach in ~\cite{sun_test-time_2020}. For remaining experiments, we therefore use the separate training strategy for TTTFlow and TTT.

\begin{figure}
    \centering
    \includegraphics[scale=0.5]{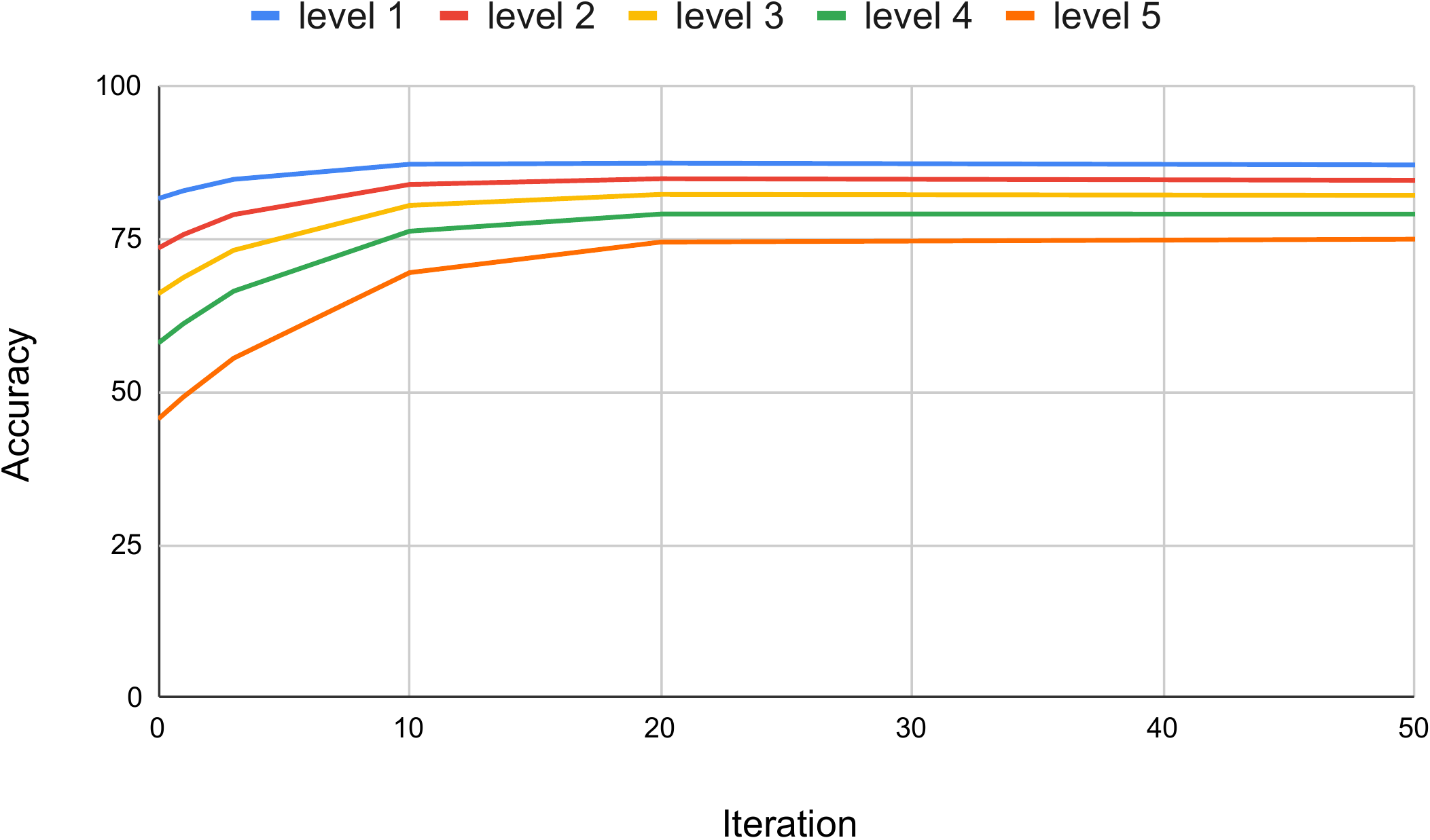}
    \caption{Evolution of accuracy over iterations of the average of each Level}
    \label{fig:AccuracyAverage}
\end{figure}

\begin{figure}
    \centering
    \includegraphics[scale=0.5]{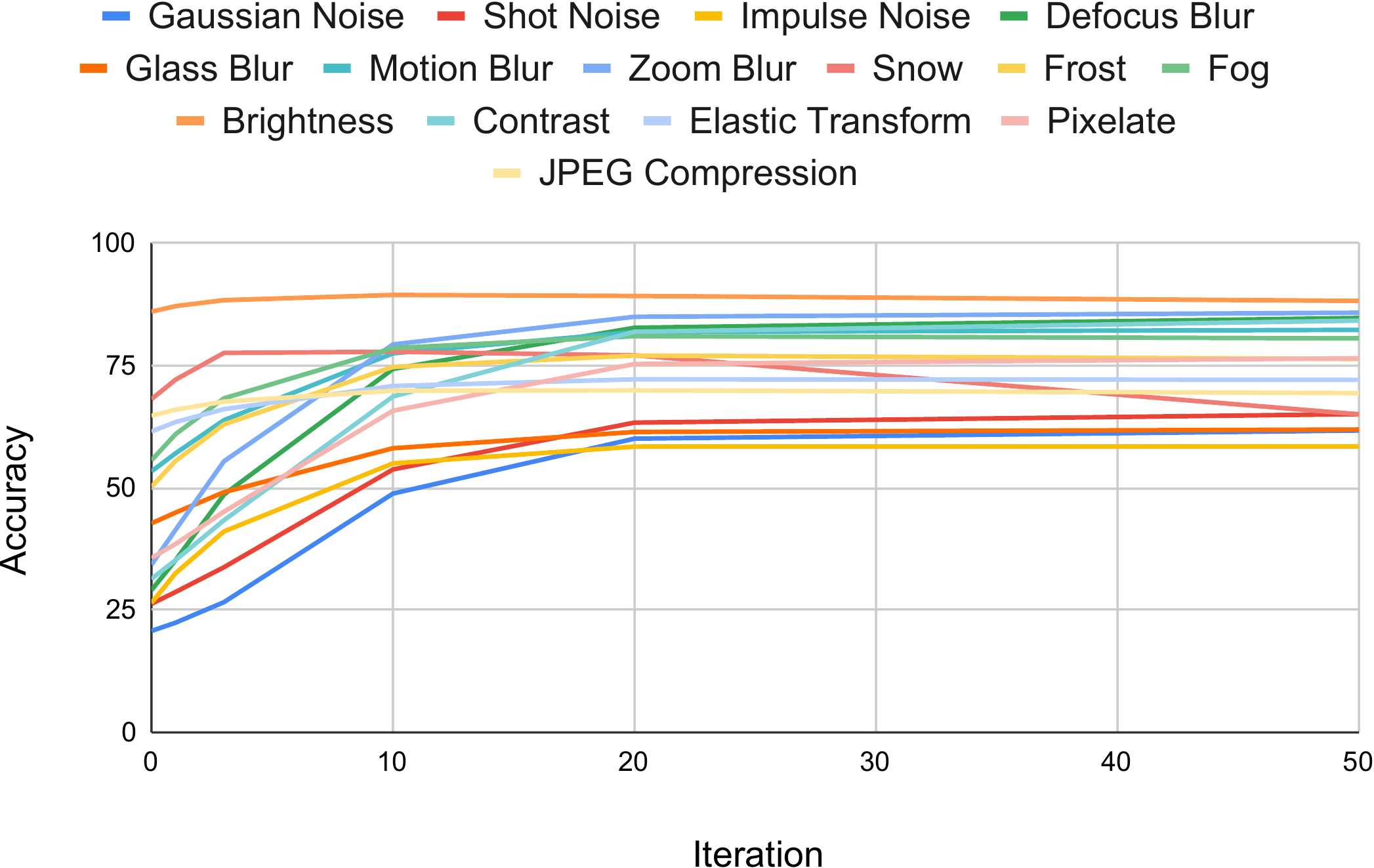}
    \caption{Evolution of accuracy for every corruption on Level 5.}
    \label{fig:AccuracyCorruptions}
\end{figure}

\mypar{Number of adaptation iterations} We compare the accuracy of TTTFlow for different iterations of adaptation at test-time.
As we can see in Fig.~\ref{fig:AccuracyAverage} and~\ref{fig:AccuracyCorruptions}, our model's accuracy typically increases monotonically with a greater number of iterations. Furthermore, in most cases, a maximum accuracy is reached after about 20 iterations. Beyond this point, performing more iterations increases runtime without any significant gain in performance. For some corruptions like Snow, which severely degrade the image, we find that performance actually drops when doing more adaptation iterations. When testing other approaches, we do not observe the same stability of performance with respect to the number of iterations. To have a fair comparison, for all methods, we thus compute the accuracy for 1, 3, 10, 20 and 50 iterations and report the maximum accuracy.

\mypar{Comparison to methods using a classifier trained with only $\Loss_{\mr{cls}}$} As shown in Table~\ref{tab:Cifar10-Clevel5}, TTTFlow achieves an average accuracy improvement of 14.54\% with respect to the pretrained ResNet50 Baseline. Significant improvements in accuracy are obtained for all corruption types except JPEG compression and Elastic transform. Moreover, TENT yielded a very low accuracy for all corruption types, due to collapsing predictions. Compared to TTT, using the same classifier trained with only $\Loss_{\mr{cls}}$, our method obtains an improvement in average accuracy of 0.84\%. These results support our hypothesis that the NF can be used to measure domain shifts and improve the extractor accordingly in an unsupervised manner.
As said in the article and in addition with these results, it confirms our hypothesis that the NF can be used to measure domain shifts and improve the extractor accordingly in an unsupervised manner.

\mypar{Comparison with TTT++ on baseline trained with $\Loss_{\mr{cls}}$ and $\Loss_{\mr{ssl}}$} As discussed earlier, the unsupervised head of TTTFlow can be placed on top of any pretrained feature extractor. In TTT++ \cite{liu2021ttt++}, the pre-training of the CNN is based on a similar Y-shaped architecture, where the secondary task is a self-supervised classification task using the contrastive loss. The joint learning process is then performed using Eq.~\ref{eq:ttt++loss} as the final loss function:

\begin{equation}
    \Loss \, = \, \Loss_{\mr{cls}} + \lambda\Loss_{\mr{ssl}}
    \label{eq:ttt++loss}
\end{equation}

where $\lambda$ is a hyperparameter. Using contrastive learning as an auxiliary task yields to a more robust network, and in consequence, a stronger feature extractor. For this reason, we propose to train the NF model using the TTT++ pre-training. As seen in Table~\ref{tab:Cifar10-Clevel5}, our method outperforms TTT++ in all but two corruption types (Impulse noise and Fog) and gives an average accuracy 2.10\% higher than this state-of-art approach.

\begin{figure*}
    \centering
    \begin{small}\setlength{\tabcolsep}{10pt}
    \begin{tabular}{cc}    
     \includegraphics[scale=0.45]{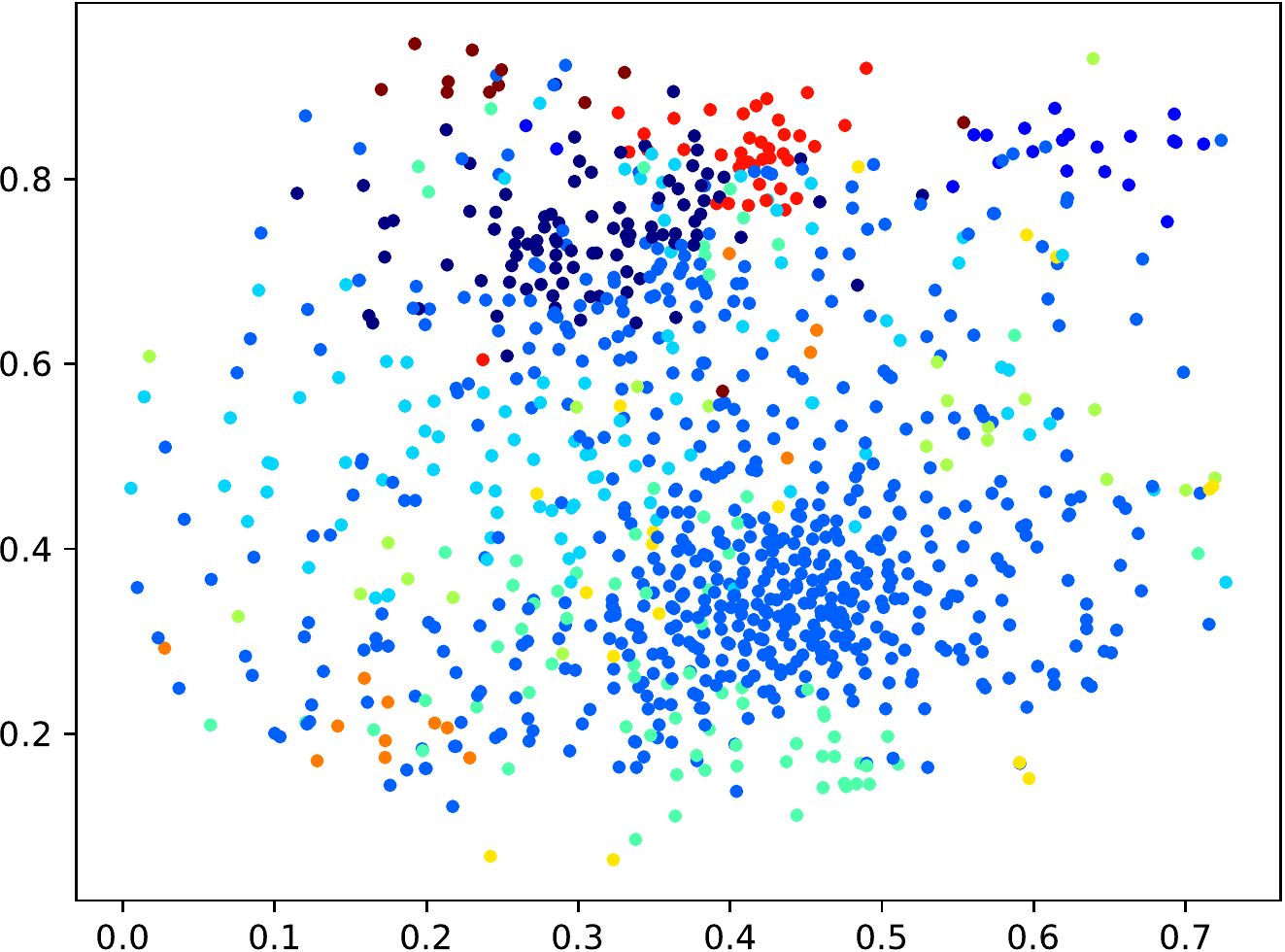} &     
    \includegraphics[scale=0.45]{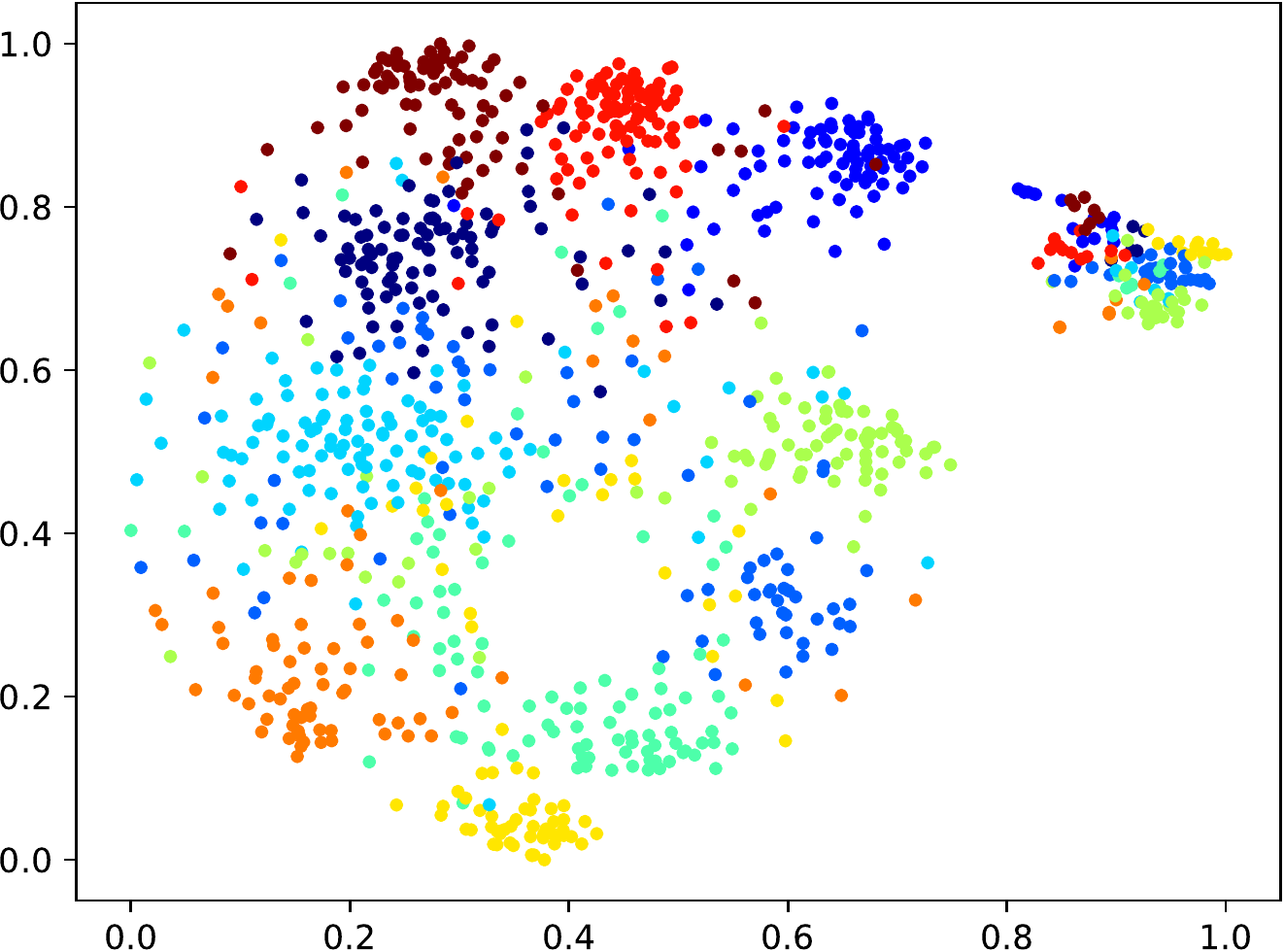}\\
    (a) & (b) \\[12pt]
    \includegraphics[scale=0.45]{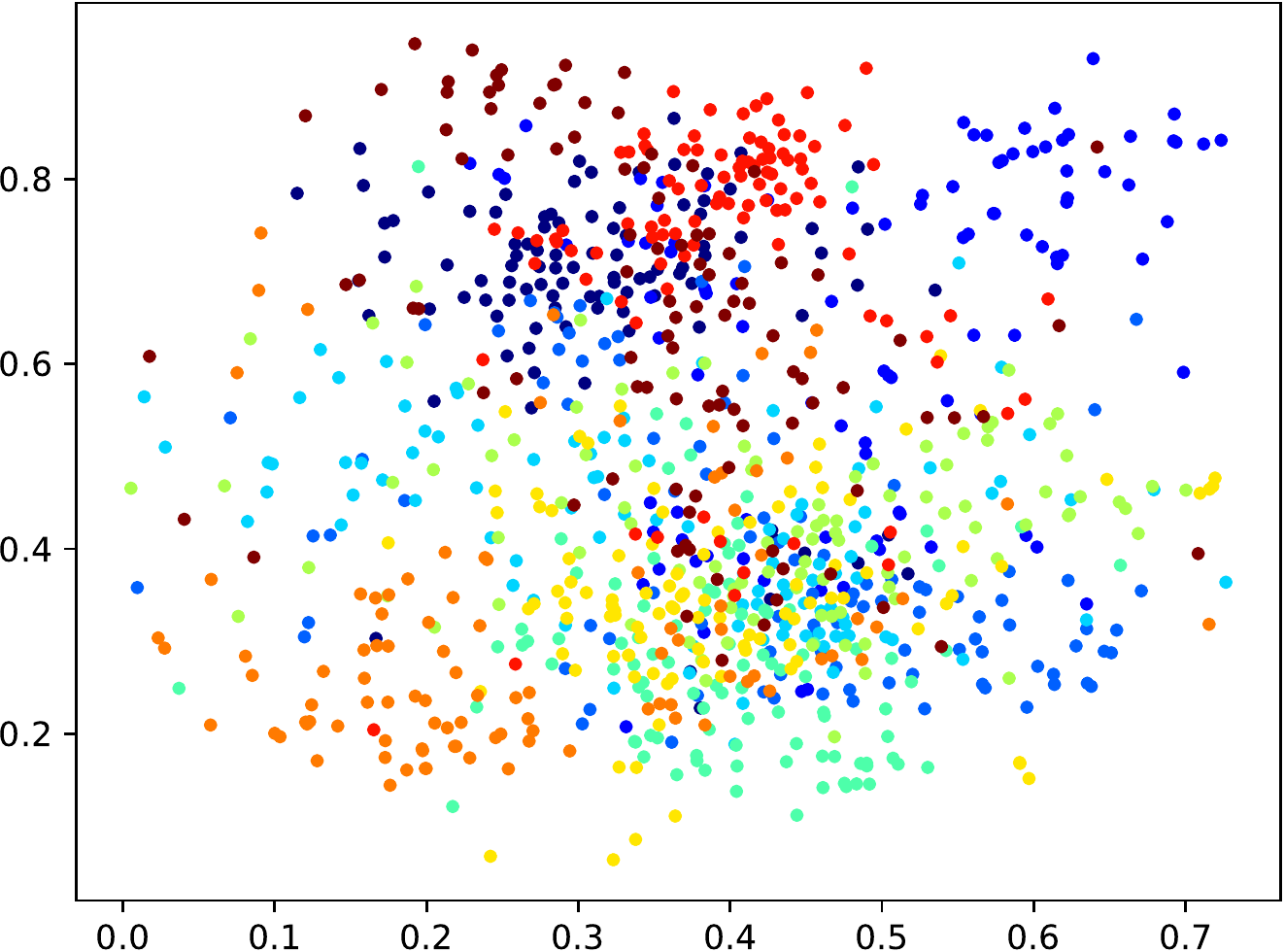} &
    \includegraphics[scale=0.45]{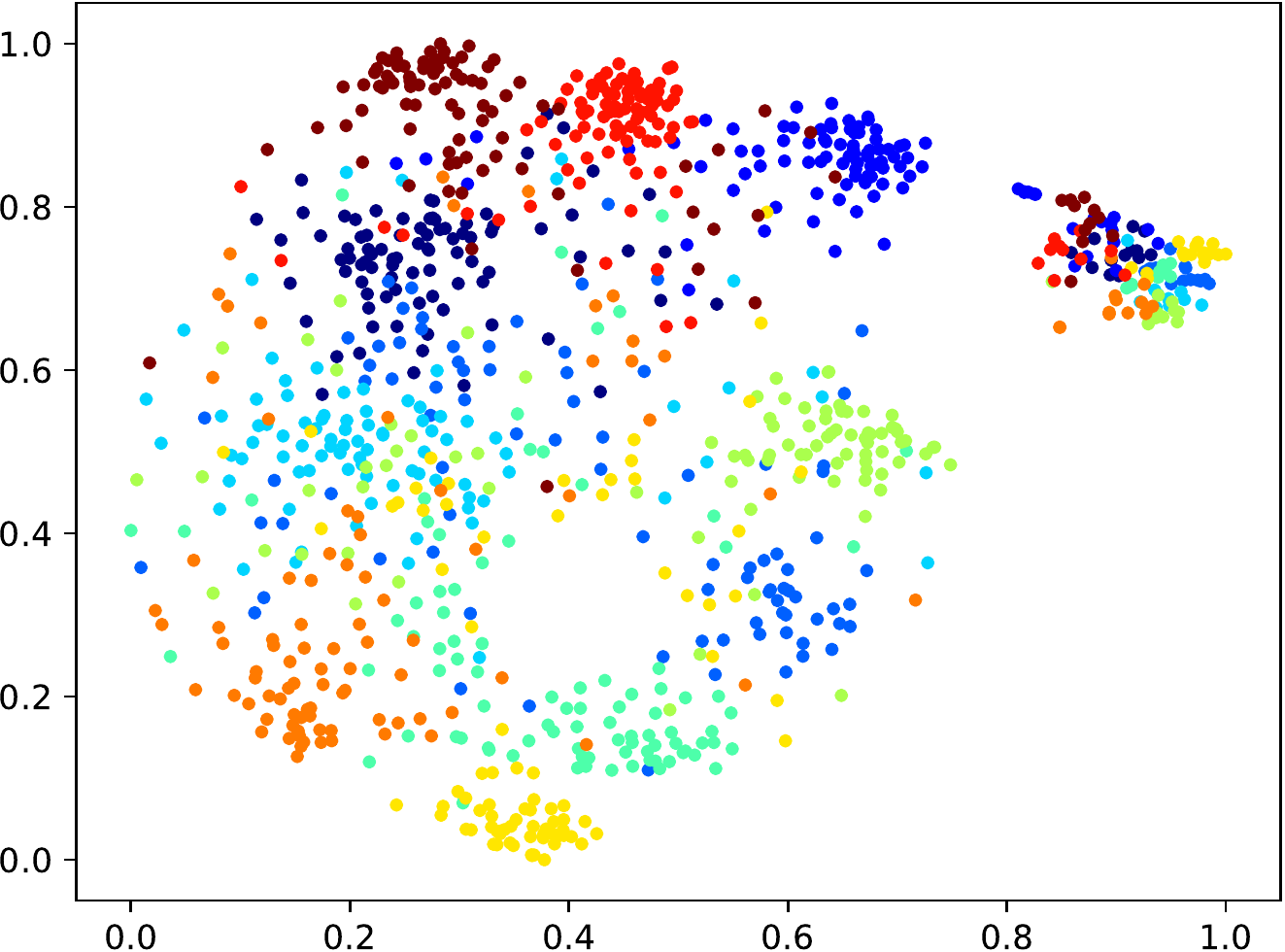}\\
    (c) & (d)
    \end{tabular}
    \end{small}
    \caption{t-SNE plots on defocus blur for the features at the output of the extractor from TTTFlow. (a) is the prediction of the model without adaptation. (b) is the prediction of the model after 50 iterations. (c) is the ground truth of the model without adaptation. (d) is the ground truth of the model after 50 iterations.}
    \label{fig:tSNEcifar10c}
\end{figure*}

\mypar{Visualization of adaptation} To visualize the result of our adaptation method, we show in Figure~\ref{fig:tSNEcifar10c} the t-SNE plots of features at the end of extractor, before and after adaptation. As can be seen, TTTFlow allows each sample to be separated for better interpretation and prediction. However, a collapse of feature vectors to the same point in space is visible in the top right of Figure~\ref{fig:tSNEcifar10c} (b) and (d). Since our NF-based method pushes representations toward the mode of the distribution, this could be a side effect of making too many iterations for adaptation. Nevertheless, in our experiments, accuracy generally remains stable and may even increase after performing many adaptation iterations.

\begin{table*}[h!]
\centering
 \caption{Accuracy (\%) on CIFAR-10-C dataset with Level 5 corruption for TTTFlow compared to ResNet50, TENT, TTT, and TTT++ with different encoders . Mean and standard deviation are reported over 5 runs}
	\label{tab:Cifar10-Clevel5}
\resizebox{\columnwidth}{!}{
        \begin{tabular}{c|cccc|cc}
        \toprule
& \multicolumn{4}{c|}{\multirow{2}{*}{Encoder trained with $\Loss_{\mr{cls}}$ only}} &  \multicolumn{2}{c}{Encoder trained} \\ 
& & & & &  \multicolumn{2}{c}{with $\Loss_{\mr{cls}}$ and $\Loss_{\mr{ssl}}$} \\ 
\cmidrule(l{5pt}r{5pt}){2-5}\cmidrule(l{5pt}r{5pt}){6-7}
& Baseline & TENT \cite{wang_tent:_2021} & TTT \cite{sun_test-time_2020} & TTTFlow & TTT++ \cite{liu2021ttt++} & TTTFlow  \\ \midrule
Gaussian Noise & 53.25 & 46.65\,\scriptsize$\pm$0.12 & 61.29\,\scriptsize$\pm$0.07 & \textbf{61.73\,\scriptsize$\pm$0.35} & 75.87\,\scriptsize$\pm$5.05 & 7\textbf{9.58\,\scriptsize$\pm$0.09} \\
Shot Noise & 57.71 & 46.31\,\scriptsize$\pm$0.25 & 64.37\,\scriptsize$\pm$0.10 & \textbf{65.08\,\scriptsize$\pm$0.14} & 77.18\,\scriptsize$\pm$1.36 & \textbf{80.20\,\scriptsize$\pm$0.03} \\
Impulse Noise & 43.79 & 37.95\,\scriptsize$\pm$0.15 & \textbf{58.97\,\scriptsize$\pm$0.20} & 58.48\,\scriptsize$\pm$0.12 & \textbf{70.47\,\scriptsize$\pm$2.18} & 67.30\,\scriptsize$\pm$0.08 \\
Defocus Blur & 51.80 & 59.77\,\scriptsize$\pm$0.29 & 83.80\,\scriptsize$\pm$0.11 & \textbf{84.75\,\scriptsize$\pm$0.17} & 86.02\,\scriptsize$\pm$1.35 & \textbf{90.96\,\scriptsize$\pm$0.06} \\
Glass Blur & 54.69 & 41.24\,\scriptsize$\pm$0.18 & 61.23\,\scriptsize$\pm$0.29 & \textbf{61.93\,\scriptsize$\pm$0.12} & 69.98\,\scriptsize$\pm$1.62 & \textbf{71.54\,\scriptsize$\pm$0.09} \\
Motion Blur & 64.97 & 56.40\,\scriptsize$\pm$0.33 & 76.86\,\scriptsize$\pm$0.13 & \textbf{82.31\,\scriptsize$\pm$0.10} & 85.93\,\scriptsize$\pm$0.24 & \textbf{85.95\,\scriptsize$\pm$0.07} \\
Zoom Blur & 61.62 & 59.23\,\scriptsize$\pm$0.35 & 84.67\,\scriptsize$\pm$0.08 & \textbf{85.82\,\scriptsize$\pm$0.17} & 88.88\,\scriptsize$\pm$0.95 & \textbf{91.90\,\scriptsize$\pm$0.05} \\
Snow & 74.12 & 55.93\,\scriptsize$\pm$0.21 & 75.63\,\scriptsize$\pm$0.1 & \textbf{77.84\,\scriptsize$\pm$0.19} & 82.24\,\scriptsize$\pm$1.69 & \textbf{84.28\,\scriptsize$\pm$0.12} \\
Frost & 67.98 & 46.44\,\scriptsize$\pm$0.20 & \textbf{77.17\,\scriptsize$\pm$0.17} & 77.05\,\scriptsize$\pm$0.10 & 82.74\,\scriptsize$\pm$1.63 & \textbf{85.88\,\scriptsize$\pm$0.05} \\
Fog & 63.67 & 52.70\,\scriptsize$\pm$0.20 & \textbf{81.15\,\scriptsize$\pm$0.12} & 81.02\,\scriptsize$\pm$0.25 & \textbf{84.16\,\scriptsize$\pm$0.28} & 74.02\,\scriptsize$\pm$0.05 \\
Brightness & 87.16 & 66.34\,\scriptsize$\pm$0.18 & 88.84\,\scriptsize$\pm$0.09 & \textbf{89.45\,\scriptsize$\pm$0.17} & 89.07\,\scriptsize$\pm$1.20 & \textbf{92.38\,\scriptsize$\pm$0.01} \\
Contrast & 22.89 & 49.03\,\scriptsize$\pm$0.45 & \textbf{84.79\,\scriptsize$\pm$0.12} & 84.20\,\scriptsize$\pm$0.18 & 86.60\,\scriptsize$\pm$1.39 & \textbf{92.20\,\scriptsize$\pm$0.10} \\
Elastic Transform & \textbf{76.96} & 50.27\,\scriptsize$\pm$0.36 & 72.45\,\scriptsize$\pm$0.09 & 72.20\,\scriptsize$\pm$0.24 & 78.46\,\scriptsize$\pm$1.83 & \textbf{80.47\,\scriptsize$\pm$0.08} \\
Pixelate & 48.22 & 52.52\,\scriptsize$\pm$0.25 & 74.71\,\scriptsize$\pm$0.09 & \textbf{76.50\,\scriptsize$\pm$0.13} & 82.53\,\scriptsize$\pm$2.01 & \textbf{88.84\,\scriptsize$\pm$0.05} \\
Jpeg Compression & \textbf{81.42} & 56.78\,\scriptsize$\pm$0.30 & 69.75\,\scriptsize$\pm$0.24 & 69.95\,\scriptsize$\pm$0.11 & 81.76\,\scriptsize$\pm$1.58 & \textbf{87.95\,\scriptsize$\pm$0.03} \\ \midrule
Average & 60.68 & 51.84 & 74.38 & \textbf{75.22} & 81.46 & \textbf{83.56} \\ \bottomrule
        \end{tabular}}   
\end{table*}

\subsection{Object recognition on natural domain shift}

We also evaluate TTTFlow when natural domain shift is present. For this purpose, the CIFAR-10.1 dataset \cite{Recht_2018_cifar101} is used as the second benchmark. CIFAR-10.1 consists of 2,000 images sampled from the original CIFAR-10 set with the objective of maximizing domain shift with respect to the source data. TTTFlow is once again compared with previous methods, and the standard pretrained CNN is used as baseline. Results are also compared with previous methods.

\begin{table}[h!]
\centering
    \caption{Accuracy of compared methods on the CIFAR-10.1 dataset containing natural domain shift.}
	\label{tab:Cifar10.1}
\begin{small}
        \begin{tabular}{lc}
        \toprule
\bf Method & \bf Accuracy (\%) \\ 
\midrule
Baseline & 84.70 \\
TENT~\cite{wang_tent:_2021} & 58.98\ppm0.12 \\
TTT~\cite{sun_test-time_2020} & 84.49\ppm0.15 \\ 
TTTFlow ($\Loss_{\mr{cls}}$) & \textbf{85.11\ppm0.30} \\ \midrule
TTT++~\cite{liu2021ttt++} & \textbf{88.24\ppm0.17} \\ 
 TTTFlow ($\Loss_{\mr{cls}}$ + $\Loss_{\mr{ssl}}$) & 86.49\ppm0.02 \\ \bottomrule
        \end{tabular}  
\end{small}        
\end{table}

\begin{table}[h!]
\centering
    \caption{Comparison of accuracy over iterations for TTT++~\cite{liu2021ttt++} and TTTFlow on th CIFAR-10.1 dataset.}
	\label{tab:IterationEvolve}
        \begin{tabular}{ccc}
        \toprule
\multirow[b]{2}{*}{\bf Iterations} &  \multicolumn{2}{c}{\bf Accuracy (\%)}
\\
 \cmidrule(l{5pt}r{5pt}){2-3}
& TTT++~\cite{liu2021ttt++} & TTTFlow ($\Loss_{\mr{cls}}$ + $\Loss_{\mr{ssl}}$) \\
\midrule
1 & 88.19\ppm0.09 & 86.49\ppm0.02 \\
3 & 88.24\ppm0.17 & 86.41\ppm0.13 \\
10 & 86.49\ppm0.22 & 86.36\ppm0.07 \\
20 & 85.03\ppm0.99 & 86.29\ppm0.08 \\
50 & 80.43\ppm0.34 & 86.48\ppm0.07 \\ \bottomrule
\end{tabular}
\end{table}

\begin{figure*}
    \centering
    \begin{small}\setlength{\tabcolsep}{10pt}
    \begin{tabular}{cc}    
     \includegraphics[scale=0.45]{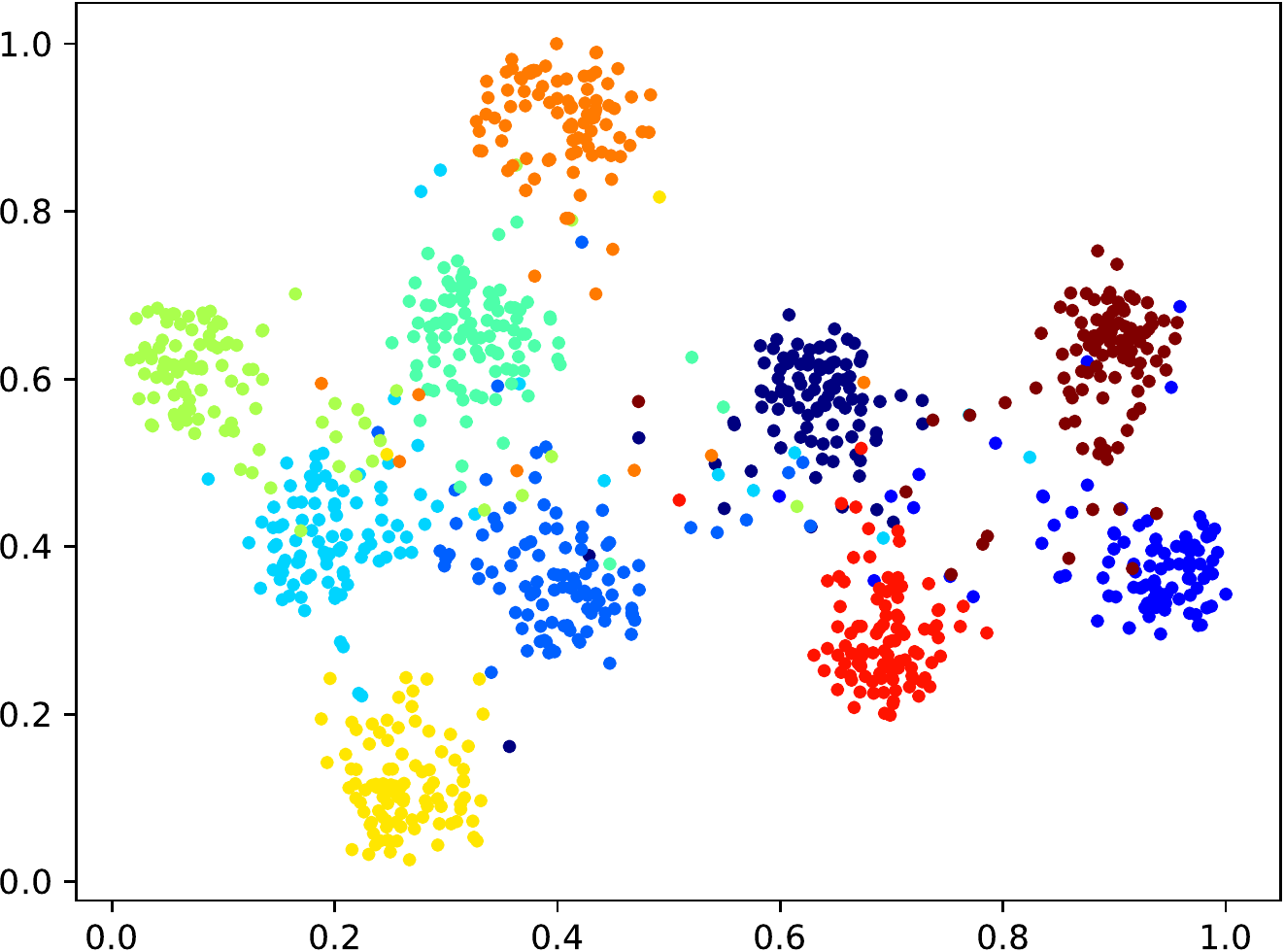} &     
    \includegraphics[scale=0.45]{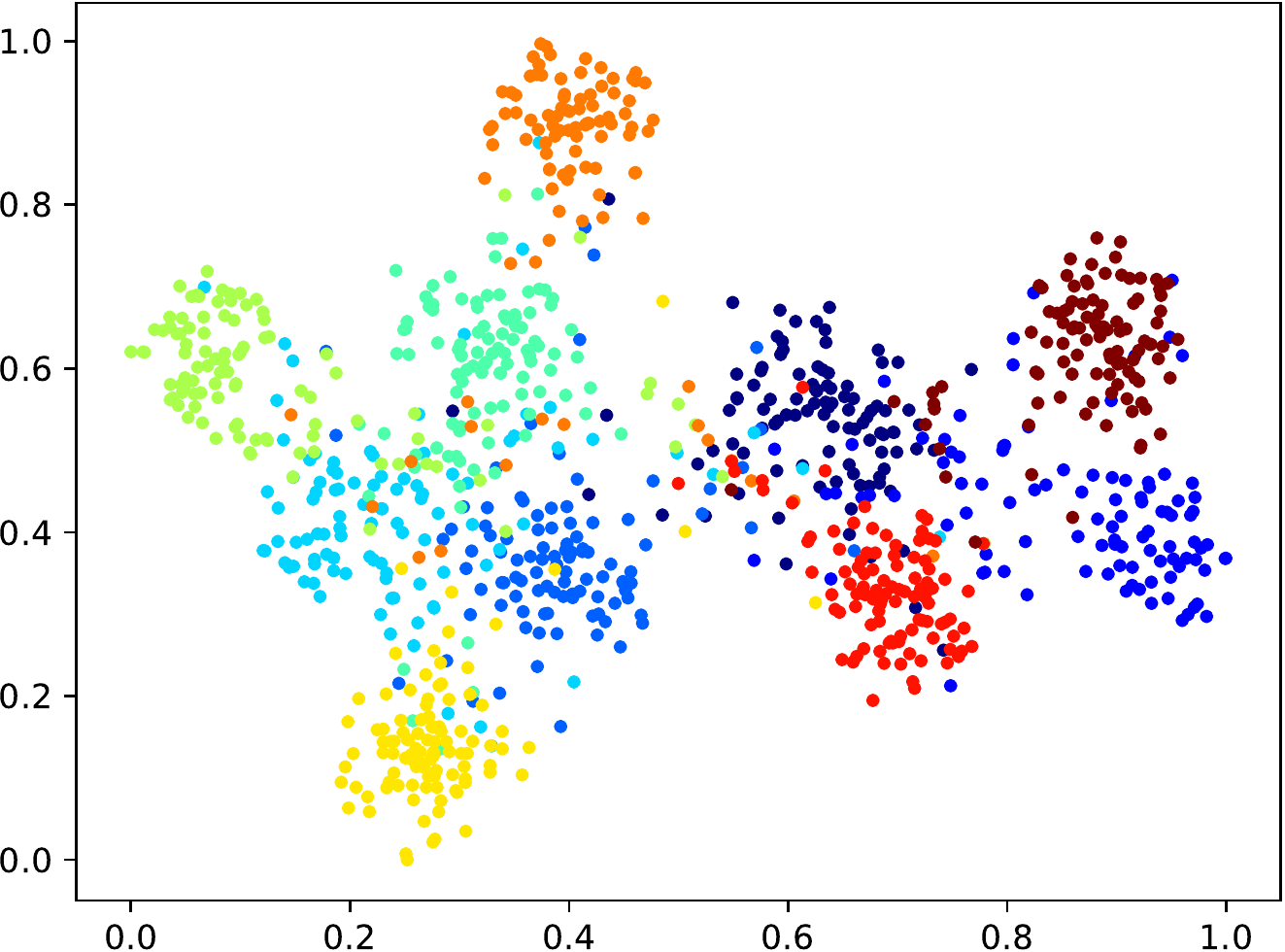}\\
    (a) & (b)
\end{tabular}
    \end{small}
    \caption{t-SNE plots for the features at the output of the extractor from TTTFlow. Comparison between CIFAR-10 and CIFAR-10.1. (a) is the ground truth of the model without adaptation for CIFAR-10. (b) is the ground truth of the model without adaptation for CIFAR-10.1.}
    \label{fig:tSNEcifar10.1}
\end{figure*}

As reported in Table~\ref{tab:Cifar10.1}, TTT++ achieves a better performance than TTTFlow in this case, with a 1.75\% improvement in accuracy. However, when looking at the accuracy for the different adaptation iteration, we find that the better performance of TTT++ only occurs for the first few iterations. Compared to our method, which remains stable, TTT++'s accuracy degrades beyond 3 iterations. The superior performance of TTT++ for this CIFAR10.1 could be explained by the nature of this dataset, in which the distribution shift is smaller compared to the corruptions found in CIFAR-10-C. The distribution shift in this dataset, which is more related to semantic content, may not be fully captured by our NF model applied on the second ResNet50 layer. As shown in the t-SNE plots of  Figure~\ref{fig:tSNEcifar10.1}, the features obtained by our model at the end of the extractor are very similar for CIFAR-10 and CIFAR-10.1, which supports that the adaptation over the iterations (Table~\ref{tab:IterationEvolve}) for TTTFlow is stable.

\section{Conclusion}

This work follows the line of former research on test-time training, which develop techniques to adapt models at test-time when distribution shifts are prevalent. To tackle some limitations of previous works, we proposed using Normalizing Flows as domain shift detector that can be plugged into the feature extractor of any pretrained architecture, and that can be trained in an unsupervised manner under maximum likelihood. 

Our method, TTTFlow, provided of substantial accuracy gains to the source model, also in comparison with the \emph{state-of-the-art} methods in test-time adaptation. Besides the practical advantage of being compatible without any model and not requiring a special joint training, it has been shown that TTTFlow can also enhance the performance of strongly trained source models, such as the one of a similar work, TTT++.

Future work includes tackling the perceived limitations of TTTFlow, which include: (a) sensitivity to the Normalizing Flow architecture, where a lower representation power could yield underfitting to the domain information, and a higher one could lead to a class collapse. (b) Depending on the type of domain shift, different layers in the encoder can be more useful to capture domain specific information, for which further studies on the effects of the chosen shared stage of the extractor are encouraged. (c) So far, TTTFlow depends on the use of batches, whilst adapting for a single sample is highly desirable. Devising a criterion to select which samples the model should adapt to would have an important impact both in performance and computational costs. 

\bibliographystyle{unsrtnat}
\bibliography{main}  






\end{document}